\documentclass{article}
\usepackage{spconf,amsmath,graphicx}

\usepackage{booktabs}
\usepackage{marvosym}

\usepackage{caption,color}
\usepackage{multirow}

\title{
Incremental Image Labeling via Iterative Refinement
}

\name{Fausto Giunchiglia$^{\star}$, Xiaolei Diao$^{\star}$
, Mayukh Bagchi$^{\star}$
}
\address{
$^{\star}$ DISI, University of Trento, Italy \\
\{fausto.giunchiglia, xiaolei.diao, mayukh.bagchi\}@unitn.it
}

\begin{document}
%
\maketitle
\begin{abstract}
Data quality is critical for multimedia tasks, while various types of systematic flaws are found in image benchmark datasets, as discussed in recent work. In particular, the existence of the semantic gap problem leads to a \textit{many-to-many mapping} between the information extracted from an image and its linguistic description. This unavoidable bias further leads to poor performance on current computer vision tasks. To address this issue, we introduce a Knowledge Representation (KR)-based methodology to provide guidelines driving the labeling process, thereby indirectly introducing intended semantics in ML models. Specifically, an iterative refinement-based annotation method is proposed to optimize data labeling by organizing objects in a classification hierarchy according to their visual properties, ensuring that they are aligned with their linguistic descriptions. Preliminary results verify the effectiveness of the proposed method.

\end{abstract}

\vspace{-0.2cm}
\section{Introduction}
\vspace{-0.3cm}
\label{S1}
Data is critical in machine learning (ML) systems, e.g., multimedia understanding,  since it is one of the core infrastructures\cite{halevy2009unreasonable}. As a data-driven science, ML is affected by data quality, which in turn impacts downstream work. Sambasivan et al. \cite{sambasivan2021everyone} explore the downstream effects of data problems, and find data cascades usually exist in tasks that underestimate the quality of data. Thus, early intervention, especially in the data collection and labeling process which is the one step that impacts the downstream most, is of vital importance to improve the performance of various tasks.

Recent work reports various types of systematic flaws in the development of object recognition benchmark datasets. For example, Dimitris et.al. provide an extensive analysis of the labeling mistakes in ImageNet \cite{IMAGENET-2009}, and point out that noisy data collection pipelines lead to systematic misalignment between generated benchmarks and real-world tasks. Shreya et.al. \cite{2017-GeoD} critically analyze the \emph{geodiversity} of ImageNet as well as OpenImages \cite{openimages} and show that they exhibit \emph{Amerocentric} and \emph{Eurocentric} representation bias. Lucas et. al. \cite{2020-AnntPipelineError} describe the bias inherent in the annotation pipeline via which ImageNet was constructed. Terrance et. al. \cite{2019-CVPR} report significant discrepancies in the classification accuracy of six \emph{`in production'} object recognition systems, and ties the discrepancies to diversity in \emph{socio-economic status}, \emph{culture} and \emph{language} from where the images were sourced. 

Our intuition is that these flaws are grounded in the way language and perception interact. The key observation is that there is a misalignment between what computer vision (CV) systems perceive from media and the words that humans use to describe the same sources. 
Specifically, current datasets utilize words or phrases to label images. For example, all category labels in ImageNet are words/phrases taken from WordNet, and we call them \textit{lexical labels}.
During the labeling process, the use of such \textit{lexical labels} will make a significant impact on the quality of the constructed dataset: the ground truth of the dataset will be directly affected by user experience, and users with different backgrounds may give inconsistent labeling results since they have different understandings to the same \textit{lexical labels}  and images \cite{giunchiglia2022visual}. 
This problem has been identified as \textit{Semantic Gap Problem} (SGP) \cite{SGP-2000}, where it was crystallized in \cite{SNCS-2021} as the fact that there is a \emph{many-to-many mapping} between the information extracted from the visual data and their possible contextual linguistic interpretations. The SGP is actually a consequence of the fact that the linguistic descriptions of an image are subjective and context-dependent. An interesting example is different objects icebergs and ice cubes are both regarded as ``ice" in Figure 1. The SGP exists not only between different people but also appears in the same person in different scenarios.

\begin{figure}[t]
	\centering
	\includegraphics[width=0.75\linewidth]{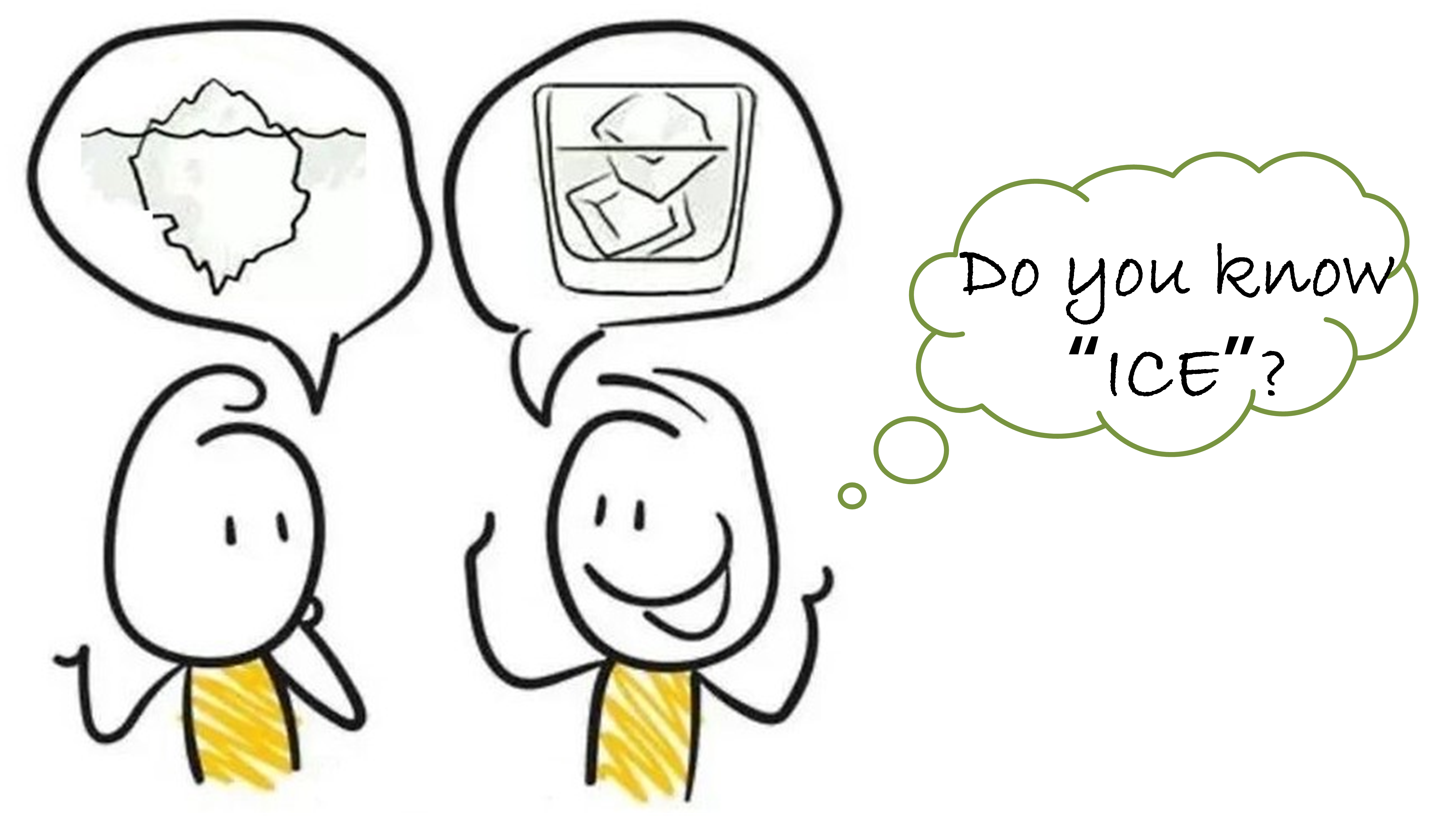}
	\vspace{-0.2cm}
    \caption{An interesting case of the semantic gap problem.}
    \vspace{-0.6cm}
    \label{fig:1}
\end{figure}

To address the above issues, we introduce a knowledge representation (KR) theory \cite{2016-FOIS} and approach into the process of data collection and labeling. We propose an image labeling process based on iterative refinement, aiming to generate high-quality ground truth datasets. The intuition stems from the fact that the current labeling process is largely unspecified, leaving much freedom to annotator subjective judgments, who can then select among the many SGP mappings. In this paper, we focus on alleviating two types of ambiguity caused by SGP, i.e., \textit{object ambiguity} and \textit{visual ambiguity}. The idea is to apply KR-base methodology to introduce the human experience to provide guidance that drives the labeling process, thereby indirectly introducing \textit{intended semantics} in the ML model, codified in a natural language format. During this process, machines and humans will refine the ground truth through iterative interaction and collaboration to obtain higher-quality datasets. We organize the labeling process based on a two-step labeling strategy, where each step is responsible for a specific aspect of the SGP many-to-many mapping, as (i) localizing objects in an image to eliminate a possible source of \textit{object ambiguity}.(ii) identifying \textit{visual properties} used to characterise objects rather than \textit{lexical labels} to eliminate a possible source of \textit{visual ambiguity}.

The contributions of our work are summarized as follows:
\begin{itemize}
\vspace{-0.2cm}
    \item We introduce a KR-based methodology to guide image labeling and constitute a paradigm shift by attempting to integrate CV with KR.
    \vspace{-0.3cm}
    \item During dataset labeling, we refine the ground truth of the dataset by introducing humans into an iterative process, and improve dataset quality by asking humans to provide feedback and supervision.
    \vspace{-0.3cm}
    \item Preliminary experiment results demonstrate that the datasets constructed based on our proposed methodology keep a higher inter-annotator agreement. 
    \vspace{-0.3cm}
\end{itemize}

The remainder of the paper is organized as follows. Sec. \ref{S2} the mislabeling types that arise in current image datasets. We illustrated the proposed labeling strategy in Sec. \ref{S3} and detail the labeling process in Sec. \ref{S4}. Preliminary results are given in Sec. \ref{S5}. Finally, Sec. \ref{S6} concludes the paper.

\vspace{-0.2cm}
\section{Types of mislabeling}
\begin{figure}[!t]
\includegraphics[width=1\linewidth]{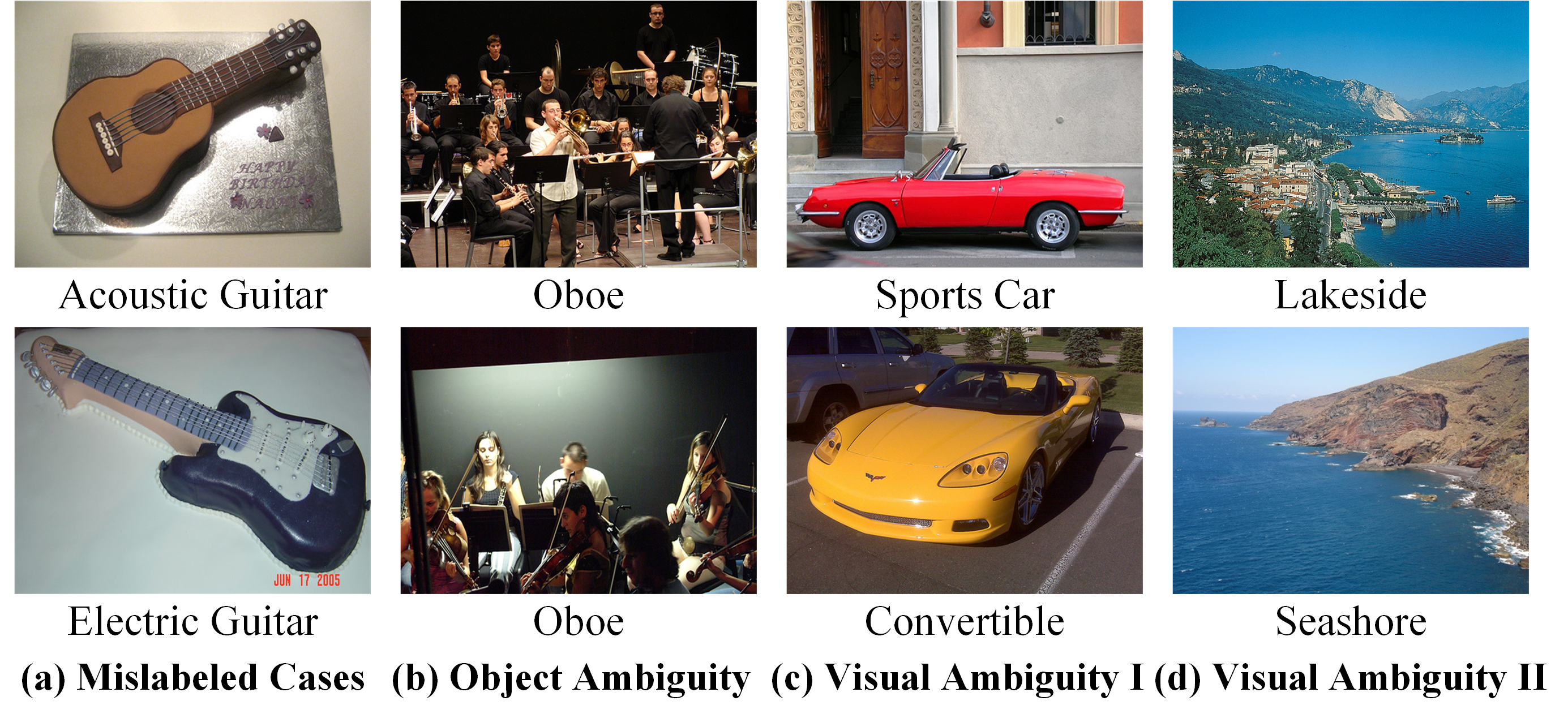}
\vspace{-0.2cm}
\caption{Mislabeled samples of different types, where the labels are from ImageNet.}
\centering
\label{example}
\vspace{-0.6cm}
\end{figure}

\vspace{-0.3cm}
\label{S2}
We take ImageNet as an example to explore the ground truth in object recognition datasets and analyze the mislabeling in it. 
We divide the mislabeling cases into two types, i.e., object ambiguity and visual ambiguity. We give some examples of the above two types in Fig.~\ref{example}, and analyze the details below.
Note that in this process, we ignore the ``simple" mistakes caused by the carelessness of the annotators since they are easily identifiable, such as an image labeled by ImageNet as \emph{`acoustic guitar'} is a `fake' guitar shaped on a birthday cake, as the two images shown in Fig.~\ref{example}(a).

\noindent \textbf{Object Ambiguity.} This type arises from the presence of multiple objects in an image. It occurs when there is an \emph {systematic incongruence} between the ImageNet label and the label of the most likely main object, as deemed by humans. An example is given in Fig.~\ref{example}(b), which is labeled \emph{`oboe'} in ImageNet but labeled \emph{`orchestral'} by humans. 
In fact, empirical evidence from cognitive psychology \cite{ROSCH-1976} suggests that humans select the main object by perceptual attributes (\emph{stimuli}) when looking to multi-object images, since it is usually the most visually salient. 
However, we observe that ImageNet exhibits biases for many multi-object images. The labels of these images correspond to a \emph{distinctive object} rather than the main object in the image, making it a challenge for models to extract features from the corresponding categories\cite{ICML-2020}.

\noindent \textbf{Visual Ambiguity.} When annotators label images, ambiguity arises in the understanding of visual data. This is determined by the background and experience of different annotators. Images are visually polysemic when their visual \emph{``semantics are described only partially"} \cite{SGP-2000}. Interpretations of images are not unique, so an image may be described with different labels by different annotators. As shown in Fig.~\ref{example}(c), an example is labeled by different annotators as \emph{`sports car'} and \emph{`convertible'}, which are two same-level classes in ImageNet. Furthermore, the two similar images in Fig.~\ref{example}(d), labeled \emph{`seashore'} and \emph{`lakeside} in ImageNet, are also representative examples that confuse the annotators. 
Such confusing class pairs are caused by \emph{design factors} of ImageNet, namely labeled only based on \textit{lexical labels}. As a result, choosing disjoint labels grounded in linguistic properties is insufficient for humans to visually disambiguate confusing class pairs in the face of potential overlapping image distributions.

Due to the semantic gap problem described above, even aside from occasional annotator errors, the resulting dataset may not accurately capture the ground truth, which seriously affects the quality of the dataset.

\vspace{-0.2cm}
\section{Two-step Labeling strategy}
\vspace{-0.3cm}
\label{S3}
In this section, we introduce a two-step labeling strategy and analyze the reasons that this strategy is able to effectively bridge the mislabeling caused by the SGP. Before that, we summarize the characteristics of \textit{good images} in a dataset by analyzing images in ImageNet with high inter-annotators consistency. Firstly, these images are almost always those containing a single object (the \emph{`main object'}, as called in \cite{ICML-2020}). Secondly, these images are less noisy, in the sense that they have the minimum influence of confounding variables such as occlusion and clutter distorting them. Thirdly, all of these images are captured from an optimal viewpoint leading to clear visibility of their defining visual characteristic.

The proposed two-step labeling strategy is as follows:
\begin{itemize}
\vspace{-0.2cm} 
    \item \textit{S1: Object localization.} Identify object(s) in an image, thus eliminating the possible \textit{object ambiguity}.
\vspace{-0.2cm} 
    \item \textit{S2: Visual classification.} Identify the \textit{visual properties} used to characterize objects, thus eliminating a possible source of \textit{visual ambiguity}. 
\vspace{-0.2cm}  
\end{itemize}

Note that such two steps are not splits of currently popular task \textit{object detection}, but are designed based on different purposes. Localization is inherently visual, while classification is inherently \emph{semantics-intensive} and is performed after object localization. 
Based on this distinction, and following the empirically validated theory of \emph{teleosmantics} \cite{2016-FOIS,millikan2020neuroscience}, we distinguish two types of concepts, \emph{substance concept} (SC) and \emph{Category Concept} (CC) \cite{millikan2005}, to represent objects. As a result of object localization, SC can be seen as a visual representation of a single object, which is in turn amenable for visual classification. As the representations of entity concept-oriented language descriptions, CC is described by natural language descriptions (i.e., lexical labels) in the current datasets.
In this process, given a continuous feed of images, the goal is to organize objects into a classification hierarchy based on their pre-defined visual features.

\noindent \textbf{S1: Object Localization}. Object localization refers to activities in which all objects in an image are localized (but not identified), for instance via bounding polygons \cite{ILSVRC}. This step aims to eliminate the possibility of \textit{object ambiguity} by identifying and extracting relevant objects in the images, as far as possible to ensure that they satisfy the basic characteristics of a \textit{good images}.
During this process, SC is represented by localizing objects in multiple images, which does not distinguish between individuals (e.g., \emph{`oboe\#123'}) and classes (e.g., \emph{`oboe'}). Meanwhile, the key to the continuous localization of a substance is grounded in its internal \emph{causal factor} \cite{2016-FOIS} which is incrementally manifested and extracted as perceivable visual properties.

\noindent \textbf{S2: Visual Classification.} The goal of step S2 is to identify the visual properties used to characterize objects, thus eliminating possible sources of \textit{visual ambiguity}. We predefine two sets of visual properties for objects, called \emph{visual genus} and \emph{visual differentia}, and build \emph{visual subsumption hierarchies} by perceiving visual properties from the new images. 
Specifically, \emph{visual genus} is a set of visual properties shared across different objects, while \emph{Visual differentia} refers to another set of visual properties different from \emph{visual genus}, which are exploited to distinguish objects within the same genus. For example, the \emph{visual genus} of the class \emph{`lobster'} is ``marine creatures with carapace", and one pre-eminent way of visually classifying further can be based on the \emph{visual differentia} ``the presence and shape of claws" with its different instantiations, i.e. ``without claws" (Spiny lobster) and ``with large tender claws"(American lobster).

One observation is that the many-to-many mapping of the SGP still occurs in the \textit{good image} category, which is caused exactly by the choice of different differentia for the same genus. That is why the classification based on category labels suffers from the SGP many-to-many mapping. SGP appears when annotators, even the same annotator, implicitly apply a different visual differentia to two images of the same object when selecting category labels based on their personal experience. Based on the above considerations, we make three fundamental assumptions in S2:
\begin{itemize}
\vspace{-0.2cm}
    \item The object hierarchy is built based on the visual genus and differentia of objects rather than category labels.
\vspace{-0.2cm}
    \item The visual properties on which the differentia is computed are consistent across all objects in that category.  
\vspace{-0.6cm}
    \item The visual properties used to compute the differentia is consistent with the modeling decisions that are taken linguistically, i.e., with the genus and differentia defined by the gloss of the corresponding WordNet class.
\vspace{-0.2cm}
\end{itemize}

\vspace{-0.2cm}
Note that our approach differs significantly from mainstream CV, especially from the way ground truth datasets have been generated so far. Furthermore, the visual classification proceeds by a (successive) selection of visual differentia assume in an egocentric setting. In other words, how to select the visual differentia completely depend on the background, point-of-view, experience, and purpose of the annotators. In practice, the selection of what we define as visual differentia depends on the highly egocentric differentiation of \emph{affordances} \cite{1977-Gibson}, with the guidance of cognitive psychology \cite{1989-Palmer}, in particular tends to visual properties that correspond to object functions (e.g., the visual property ``a pair of joined reeds that vibrate together" for an oboe denotes the function of playing them to produce sound). 

\vspace{-0.2cm}
\section{The iterative labeling process}
\vspace{-0.3cm}
\label{S4}
In this section, we detail the image labeling methodology. This is an iterative refinement process, which includes three loops, namely the top-level loop, the vertical loop, and the horizontal loop. The first step object localization is applied in the top-level loop, and the second step visual classification produced by the vertical loop and horizontal loop.

\noindent \textbf{The Top-level Loop.} The top-level loop is designed to continuously offer \textit{good images} for the iterative labeling method. In this process, the \textit{object localization} step is introduced to avoid \textit{object ambiguity}. An object localization model \cite{redmon2018yolov3} is introduced to automatically locate objects by machine in an image, and crop the image based on the coordinates of objects, aiming to obtain images with a single object. As a result, we can obtain multiple single-object images from one multi-object image. Note that, although the square bounding box sometimes contains parts of other objects in the cropped image, the main object of these images will be more defined than the original image, thus, we treat the cropped image as a single-object image. Then, all images are input to the following vertical loop and horizontal loop one by one for labeling.

\noindent \textbf{The Vertical Loop.} The goal of the vertical loop is to iteratively refine the label of the input image through layer-by-layer computation to label it precisely. There is also a hierarchy that is input at the same time as the image, which will be gradually enriched by adding nodes and samples in the current iterative process. When inputting a new object, the similarity is applied to compare with the categories already stored in the hierarchy, and the category with the highest similarity is regarded as the ``candidate". Next, humans join this loop to determine whether the object has the common \textit{visual genus} as the ``candidate". If the answer is ``False", proceed to compare this object to other categories at the same layer with ``candidate" in the hierarchy, and enter the horizontal loop until ``True" is obtained. None of ``True" obtained means the object does not share a \textit{visual genus} with any categories, then it will be labeled as a new category and stored in the hierarchy. During this process, the machine is responsible for recommending initial labeled options and performing the method, while humans take charge of determining the \textit{visual genus} to decide whether further labeling refinement is needed.

\noindent \textbf{The Horizontal Loop. } The goal of the horizontal loop is to label the most refined category for the input object by comparing them within a domain category. It is triggered by the vertical loop and compares the input object with the subcategories of the candidate (if the candidate category has no subcategories, the input object is labeled as the candidate category). The process starts from the subcategories with the highest similarity of the object and asks humans whether they have \textit{visual differentia}. If the answer is ``False", it means that they belong to the same category, and the input object is labeled as the current subcategory; if the answer is ``True", the same comparison is proceed with the next subcategory. If all subcategories in the candidate has \textit{visual differentia} from the input image, it means the object does not belong to any subcategory, and it is labeled as a new category and added to the hierarchy as a new subcategory of the candidate. In the process, we complete the labeling and enrich the hierarchy at the same time. This incremented hierarchy will also continue to be utilized in the next top-level loop.

\noindent \textbf{Observation and Analysis} There are two important observations. Firstly, the fact that, though we have a detailed set of canonical principles for ensuring the \textit{visual subsumption hierarchy} to be ontologically thorough, the task becomes particularly critical due to the tradeoff between the appropriate vertical and horizontal choice in uniquely classifying an object. The choice must be guided by the specific object recognition task that must be performed by the model trained using the dataset generated. In other words, there cannot be a \textit{fits-it-all} dataset. Instead, we envisage a future where this methodology will allow the construction of datasets with clear and precisely specified \textit{semantic} properties, which will then be introduced in the ML models by using them for training.
Secondly, as a consequence of the first observation, human supervision can often be necessary for determining the exact (succession of) differentia in sync with the egocentric hierarchy in the mind of the user. 
The key observation underlying both observations, also factoring in other phases, is that the faceted classification process, while (of course) not eliminating human subjectivity, does provide the guidelines for \emph{enforcing a one-to-one mapping} between visual and linguistic properties. Overall, although the proposed iterative refinement process of image labeling still suffers from human subjectivity, it does provide the guidelines for \emph{enforcing a one-to-one mapping} between visual and linguistic properties.

\vspace{-0.2cm}
\section{Preliminary Results}
\vspace{-0.3cm}
\label{S5}
\begin{table}[!t]
\setlength{\abovecaptionskip}{3pt}%
\setlength{\belowcaptionskip}{5pt}%
\caption{The image labeling results by two annotators, where ``1, 1\_1, ..." represent nine different categories, respectively, and the ``Alpha" is Krippendorff’s alpha measure.}
\label{tab:1}
	\resizebox{1\linewidth}{!}{%
\begin{tabular}{c|ccccccccc|c}
\hline
\textbf{Categories} & \textbf{1}  & \textbf{1\_1} & \textbf{1\_1\_1} & \textbf{1\_1\_1\_1} & \textbf{1\_1\_1\_2} & \textbf{1\_1\_2} & \textbf{1\_1\_3} & \textbf{1\_2} & \textbf{1\_3} & \multicolumn{1}{l}{Alpha} \\ \hline
Expert1    & 17 & 42   & 21      & 21         & 22         & 13      & 12      & 33   & 10   & \multirow{2}{*}{0.9832}                    \\
Expert2    & 17 & 42   & 20      & 22         & 22         & 13      & 12      & 33   & 10   &                                            \\ \hline
\end{tabular}}
\vspace{-0.2cm}
\end{table}

\noindent \textbf{Inter-annotator agreement.} To evaluate the inter-annotator agreement, we collect 191 musical instrument images of nine categories from ImageNet and invite two annotators to label these images based on our proposed iterative refinement process. After finishing labeling, the number of images for each category is shown in Table \ref{tab:1}. We introduce Krippendorff's alpha\cite{hughes2021krippendorffsalpha} for agreement measure. Results as high as 0.9832 demonstrate near-perfect agreement between two annotators, which verifies the reliability of our method.

\begin{table}[t]
\setlength{\abovecaptionskip}{3pt}%
\setlength{\belowcaptionskip}{10pt}%
\caption{Classification results on two datasets.}
\label{tab:2}
	\resizebox{1\linewidth}{!}{%
\begin{tabular}{c|ccccc}
\hline
\multirow{2}{*}{\textbf{Methods}} & \multicolumn{5}{c}{\textbf{Accuracy}}                 \\ \cline{2-6} 
                                  & \textbf{VGG} \cite{vgg} & \textbf{GoogleNet} \cite{googlenet}   & \textbf{ResNet} \cite{resnet}   & \textbf{RAN} \cite{ran}  & \textbf{SENets} \cite{senet}    \\ \hline
ImageNet (subset)                            & 0.699 & 0.727     & 0.538    & 0.706                       & 0.734  \\
Refine dataset (ours)                          & 0.762 & 0.825     & 0.741    & 0.790                       & 0.804  \\ \hline
\end{tabular}}
\vspace{-0.2cm}
\end{table}

\noindent \textbf{Machine Learning Experiment.} In this experiment, we used two different datasets to train five classic ML methods. ImageNet (subset) refers to a subset containing nine musical instrument categories collected from ImageNet with the original labels. The refined dataset (ours) is labeled by our proposed iterative refinement strategy with the same images. During training, we keep all parameters consistent and use the same test set. The results of object recognition are shown in Table \ref{tab:2}. It can be found that the accuracy of the same methods is significantly improved when trained on our dataset. These results confirm that our dataset has higher data quality, and verifies the validity of the proposed methodology. 

\vspace{-0.2cm}
\section{Conclusion}
\vspace{-0.3cm}
\label{S6}

In this paper, we propose a general iterative refinement process based on KR methodology to generate high-quality ground truth image datasets, aiming to overcome the limitations imposed by SGP on current labeling processes. In the future, we will focus on the construction of large benchmark datasets by extending the data labeling methodology.

\section{Acknowledgments}
\vspace{-0.3cm}
This research has received funding from the European Union's Horizon 2020 FET Proactive project ``WeNet – The Internet of us”, grant agreement No 823783.

\clearpage
\bibliographystyle{IEEEbib}
\bibliography{icassp_ref}

\end{document}